%% file: main.tex
\documentclass[10pt,twocolumn,letterpaper]{article}

\usepackage{iccv}
\usepackage{times}
\usepackage{epsfig}
\usepackage{graphicx}
\usepackage{amsmath}
\usepackage{amssymb}
\usepackage{gensymb}

\usepackage{enumitem}
\usepackage{booktabs}
\usepackage{multirow}
\usepackage{graphicx}
\usepackage{caption}
\usepackage{subcaption}
\usepackage[ruled,linesnumbered]{algorithm2e}
\usepackage{rotating}
\usepackage[table]{xcolor}

\usepackage[symbol*]{footmisc}
\usepackage{pifont}
\usepackage{authblk}
\usepackage[export]{adjustbox}

\newcommand{\cmark}{\ding{61}}%
\newcommand{\zmark}{\ding{67}}%

\newcommand{\avgunder}{\mathop{\mathrm{avg}}\limits}
\newcommand{\norm}[1]{\left\lVert#1\right\rVert}

\usepackage[breaklinks=true,bookmarks=false]{hyperref}

\iccvfinalcopy % *** Uncomment this line for the final submission

\def\iccvPaperID{2} % *** Enter the ICCV Paper ID here

\ificcvfinal\fi
\begin{document}

%%%%%%%%% TITLE
\title{HomebrewedDB: RGB-D Dataset for 6D Pose Estimation of 3D Objects}

\author[\zmark,\cmark]{Roman Kaskman}
\author[\zmark,\cmark]{Sergey Zakharov}
\author[\zmark,\cmark]{Ivan Shugurov}
\author[\zmark,\cmark]{Slobodan Ilic}
\affil[\zmark]{ Technical University of Munich, Germany\quad \cmark~Siemens Corporate Technology, Germany 
{\normalsize \tt{\{roman.kaskman, sergey.zakharov, ivan.shugurov\}@tum.de}}, {\normalsize \tt{\, slobodan.ilic@siemens.com}}}

\maketitle
%\thispagestyle{empty}

%%%%%%%%% ABSTRACT
\begin{abstract}
Among the most important prerequisites for creating and evaluating 6D object pose detectors are datasets with labeled 6D poses. With the advent of deep learning, demand for such datasets is growing continuously. Despite the fact that some of exist, they are scarce and typically have restricted setups, such as a single object per sequence, or they focus on specific object types, such as textureless industrial parts. Besides, two significant components are often ignored: training using only available 3D models instead of real data and scalability, i.e. training one method to detect all objects rather than training one detector per object. Other challenges, such as occlusions, changing light conditions and changes in object appearance, as well precisely defined benchmarks are either not present or are scattered among different datasets. 

In this paper we present a dataset for 6D pose estimation that covers the above-mentioned challenges, mainly targeting training from 3D models (both textured and textureless), scalability, occlusions, and changes in light conditions and object appearance. The dataset features 33 objects (17 toy, 8 household and 8 industry-relevant objects) over 13 scenes of various difficulty. We also present a set of benchmarks to test various desired detector properties, particularly focusing on scalability with respect to the number of objects and resistance to changing light conditions, occlusions and clutter. We also set a baseline for the presented benchmarks using a state-of-the-art DPOD detector. Considering the difficulty of making such datasets, we plan to release the code allowing other researchers to extend this dataset or make their own datasets in the future. 

%Moreover, the paper presents a pipeline enabling people to generate the annotated datasets of their own, further promoting research in 6D pose detection.
\end{abstract}

%%%%%%%%% BODY TEXT
%===========================================================
\section{Introduction}
\input{content-introduction}

%-------------------------------------------------------------------------
\section{Related Datasets}
\label{sec:rw}
\input{content-related-work}

%-------------------------------------------------------------------------
\section{HomebrewedDB Dataset Creation}
\label{sec:mth}
\input{content-methodology}

%-------------------------------------------------------------------------
\section{Benchmarks and Experiments}
\label{sec:exp}

\input{content-experiments}

%-------------------------------------------------------------------------
\section{Conclusion}
\label{sec:cnc}
\input{content-conclusion}

%===========================================================

%%%%%%%%% BIBLIOGRAPHY

{\small
\bibliographystyle{ieee_fullname}
\bibliography{egbib}
}

\end{document}

%% file: content-introduction.tex
% !TeX spellcheck = en_US

% \begin{figure}[t]
%   \centering
%   \includegraphics[width=1\linewidth]{figures/teaser.png}
%   \caption{
%   \textbf{Scene examples of the HomebrewedDB dataset:} Our dataset features 13 RGB-D annotated scenes of various difficulty. \label{fig:refinement}
%   }
% \end{figure}

\begin{figure}[t]
   \begin{minipage}[b]{.495\linewidth}
	\includegraphics[width=1.0\linewidth]{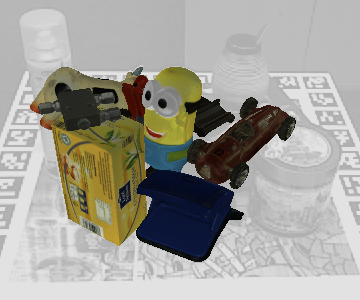}
  \end{minipage}
	\hfill
   \begin{minipage}[b]{.495\linewidth}
	\includegraphics[width=1.0\linewidth]{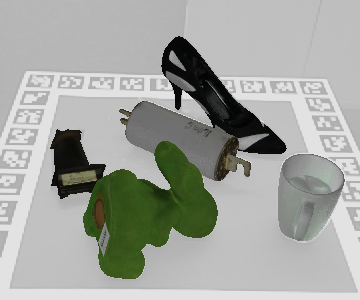}
  \end{minipage}
  
   \begin{minipage}[b]{.495\linewidth}
   \vspace{0.2em}
	\includegraphics[width=1.0\linewidth]{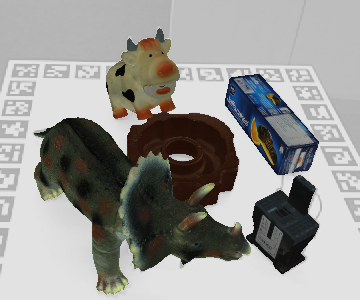}
  \end{minipage}
  \hfill
  \begin{minipage}[b]{.495\linewidth}
	\includegraphics[width=1.0\linewidth]{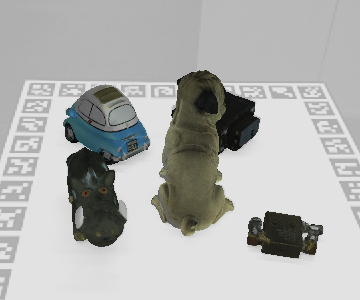}
  \end{minipage}
   \caption{
   \textbf{HomebrewedDB scene examples:} Our dataset features 13 RGB-D annotated scenes of various difficulty. The reconstructed 3D models of the objects are rendered on top of RGB images with obtained ground truth poses.} \label{fig:refinement}
   
   \vspace{-0.8em}
\end{figure}
Detection of 3D objects in images and recovery of their 6D poses is crucial for a wide range of computer vision tasks. In robotics it is essential to determine 6DoF of the object for the tasks of object grasping, manipulation and automatic assembly. Also, precise 6D poses in RGB images are of utmost importance in augmented reality (AR) applications, where overlaying 3D models on top of the real objects is critical for AR-driven assembly or repair tasks. With the rise of deep learning methods, demonstrating better performance than traditional template matching approaches or methods based on handcrafted features, it is essential to have appropriate datasets, which would encourage development and thorough evaluation of the new approaches.

A number of 6D pose object datasets exist, each focusing on one of the aspects of this challenging task. For example, datasets like T-LESS~\cite{hodan2017tless} and LineMOD~\cite{hinterstoisserModelBasedTraining2013} cover textureless and target-specific object types in particular scenarios. LineMOD contains mainly toy and household objects of distinct geometry, concentrating on object detection in cluttered environments with nonexistent or minor occlusions. It includes a total of 15 separate image sequences, however each of them features 6D pose annotations only for a single object. T-LESS is a much bigger dataset, focusing exclusively on textureless industrial objects, which exhibit strong inter-object similarities and symmetries. In T-LESS, separate training and test images are explicitly provided, while it remains unclear how to sample or produce training data from LineMOD. This ambiguity has led to wide inconsistencies in presented results, making it hard to conduct a comprehensive assessment of the developed methods. Striving for the best possible results, researchers tend to train their detectors on subsets of provided real images with ground truth poses very similar to those in a test split. This strategy inevitably leads to overfitting to a particular dataset, restricts applicability of the detectors and undermines fair comparison of various approaches.

One crucial aspect that has often been neglected by 6D pose detectors is scalability. The majority of the datasets contain a rather small number of objects. This is natural since producing large number of 3D models and scenes with annotated poses is a tedious task, especially when poses of all the objects in all frames must be available. When it comes to deep learning methods, training detectors on real data yields the best results. However, the fact that 3D models of the objects are available and training data can be synthesized by rendering them has been used in only a few studies, most notably using such detectors as SSD6D~\cite{kehlSSD6DMakingRGBbased2017b}, AAE~\cite{sundermeyerImplicit3DOrientation2018} and DPOD~\cite{zakharov2019dpod}. It is remarkable that all deep learning 6DoF object detectors trained either on real or synthetic data use a single neural network per object, in contrast to 2D object detectors, such as YOLO~\cite{redmonYouOnlyLook2015}, SSD~\cite{liuSSDSingleShot2016} or R-CNNs~\cite{girshickRichFeatureHierarchies2014a, girshickFastRcnn2015, renFasterRCNNRealTime2015, heMaskRCNN2017}, which use one network for all object classes. A major reason for this issue is the unavailability of a proper dataset with a variety of sequences and well-defined benchmarks, causing a fallback to a well-studied simplistic dataset such as LineMOD. Nevertheless, training one network per object defeats the scalability aspect that is naturally a characteristic of deep neural networks. Therefore, central aspects of our proposed dataset is its ability to test methods' scalability by introducing corresponding benchmarks and to encourage training of detectors on renderings of 3D models instead of using real data.

Our dataset may be considered to be aligned with the OCCLUSION dataset~\cite{brachmannLearning6DObject2014a}, which contains poses of all the objects present in each frame. However, instead of just a single sequence with 1214 frames, our dataset contains
13 full-circle scenes filmed with both PrimeSense Carmine 1.09 (structured light) and Microsoft Kinect 2 (time-of-light) RGB-D cameras resulting in a total of 34,830 fully annotated frames with poses for all objects in all frames. The complexity of the scenes increases from simple (several separated objects per scene) to heavily-cluttered and occluded (objects close together or on top of each other and also mixed with other objects not present as 3D models). Another aspect that is not addressed in the other datasets is strong variation of the
illumination, including not only changes in light intensity, but also in light color. Finally, driven by the fact that in industry objects can undergo severe appearance changes, we created a benchmark where the object appearance is altered compared to the one in available 3D models. 

In the following sections, besides describing the steps of the dataset creation pipeline, we also present detection and 6D pose estimation results for all the newly-defined benchmarks of one of a recently introduced methods - Dense Pose Object Detector (DPOD)~\cite{zakharov2019dpod}. This method is trained on all the objects at once or on all the objects present in the test scene on strictly synthetic renderings of provided 3D models. The aim of our experiment was not only to test scalability of the current methods but also to demonstrate that even the best-performing methods trained on real images with one network per object obtain mediocre results when trained on multiple objects. This establishes a baseline and opens a new challenge to the community related mainly to scalability and synthetic training data. Light and appearance changes are another new aspect of this challenging dataset.

%% file: content-related-work.tex
% !TeX spellcheck = en_US

Given the fact that determining ground truth 6D poses from RGB images is an ambiguous task requiring manual interventions, it is not surprising that the majority of 6D pose datasets are made with the use of RGB-D cameras. Because these cameras provide depth images aligned with color images, the task of 6D pose estimation becomes simpler and more automated. The most popular datasets for 6D pose estimation are discussed in this section.
\vspace{-1em}
\paragraph{LineMOD Dataset.} One of the most widely used 6D pose datasets is LineMOD by Hinterstoisser \etal~\cite{hinterstoisserModelBasedTraining2013}. Essentially, it contains objects embedded in cluttered scenes. It was acquired with PrimeSense Carmine RGB-D sensor and in total comprises 15 objects, two of them being symmetric. For each sequence, poses of only one object are annotated. The target objects with the existing ground truth poses are either not occluded, or are subject to very slight occlusions. The original template matching method~\cite{hinterstoisserModelBasedTraining2013}, which was published alongside the dataset, performed poorly on RGB and depth images separately, while the results obtained on RGB-D images were extremely good and still remain hard to achieve by many modern deep-learning-based methods. However, this method suffers from a couple of drawbacks: it scales poorly and cannot handle occlusions. OCCLUSION dataset was created by Brachmann \etal.~\cite{brachmannLearning6DObject2014a} in order to address the lack of occluded test data. In this extension of the original LineMOD additional manual annotations of 6D poses for all the objects in each frame were included. Due to the necessity of intensive manual labor, it was done for only a limited number of frames. Availability of RGB and depth images as well as 3D CAD models with original object colors resulted in probably the widest use of this dataset of all those targeting 6D pose estimation. Deep learning methods use RGB images from this dataset for object detection and 6D pose estimation. State-of-the-art results for real and synthetic data were achieved by the recently introduced DPOD method~\cite{zakharov2019dpod}.
\vspace{-1em}
\paragraph{T-LESS Dataset.} T-LESS~\cite{hodan2017tless} is a recent dataset that is gaining popularity. It contains 30 textureless industrial objects and 20 RGB-D scenes captured with three synchronized cameras (PrimeSense Carmine 1.09 and Kinect 2 RGB-D cameras and Canon RGB camera). The objects in this dataset have strong inter-object similarity. The acquired scenes vary from simple (less clutter and several objects per scene) to complex (heavily cluttered, with piles of objects, mimicking a typical robotic bin-picking scenario). Both hand-designed CAD models and reconstructed 3D models are included in T-LESS. Training images contain isolated objects on black backgrounds, while test images capture entire scenes with labeled 6D poses for each object in each frame. The structure of this dataset is exceptional, but due to symmetry, low texture and the industrial nature of the objects it is very challenging, which might be the reason why it has not gained in popularity at the pace it deserves. Truly inspired by T-LESS~\cite{hodan2017tless}, we prepared our dataset similar to this one in terms of structure. However, our dataset covers a wider range of object types, spanning toys, household objects as well as industrial objects. Also, we aimed to build a dataset with occlusion challenges, and that goes well beyond OCCLUSION, thereby providing many more frames and scenes for this task. %Similarly to T-LESS, since we aim at methods that will address scalability, we do not provide training images for all objects, but rather expect researchers to use reconstructed 3D models of our dataset and generate training data by rendering them. 
\vspace{-1em}
\paragraph{YCB-Video Dataset.} This dataset is very recent and came out with the PoseCNN detector~\cite{YuXiang17_PoseCNN}. Unlike T-LESS, which contains images of the scenes 
from all viewpoints, this one resembles LineMOD, containing short videos depicting several household objects in the scene. However, the large number of video sequences (92) as well as the presence of occlusions in the test scenes make this dataset quite attractive.
\vspace{-1em}
\paragraph{Other Datasets.} The following datasets have been sporadically used in some publications, but their systematic use has not been achieved. Tejani \etal~\cite{Tejani14} contains 2 textureless and 4 textured objects with 700 frames of test images. A characteristic of this dataset is that it includes multiple instances of the same object. Clutter and occlusions are moderate, which makes it not particularly challenging. Doumanoglou \etal~\cite{DoumanoglouKMK15} presented a bin-picking dataset with 183 test images and only two objects from the Tejani dataset, but  multiple instances of them. The Challenge and Willow datasets~\cite{XieSUNA13} contain a larger number of objects (176), but a relatively small number of test images (353). The TUW dataset~\cite{AldomaFV14} is similar with 17 objects in 224 test images. Datasets that are suited for robotic tasks are the Rutgers ~\cite{RennieSBS15}, the Amazon Picking Challenge~\cite{CorrellBBBCHORR16}, and the BIGBird~\cite{SinghSNAA14} datasets. Datasets addressing the light change challenge are TUD-Light and Toyota Light, both used and referenced in the benchmark paper by Hodan \etal~\cite{HodanMBKBKDVIZS18}.

Our dataset could be considered to be between OCCLUSION and T-LESS. It contains high-quality annotations, a multitude of objects and a large number of scenes and test images. %It contains very small number of training images depicting single objects seen from various viewpoints and on purpose omits this information for all objects aiming at methods trained on synthetic models. 
In contrast to the recent work of Hodan \etal~\cite{HodanMBKBKDVIZS18}, who have tried to unify 8 different datasets and have concentrated on evaluation of RGB-D methods, we are more interested in RGB methods, even though full RGB-D images are available in our dataset. With scalability at its core, we design benchmarks in which one network is trained either for all the available objects or only for the objects present in a particular scene. Additionally, we introduce scenes with severe environment changes, including changes of light colors and intensities as well as changes of object appearance. We believe that this dataset will push forward research on scalable object detection and domain adaptation.

%% file: content-methodology.tex
% !TeX spellcheck = en_US

\begin{figure*}[t!]
	\centering
	\includegraphics[width=1.0\linewidth]{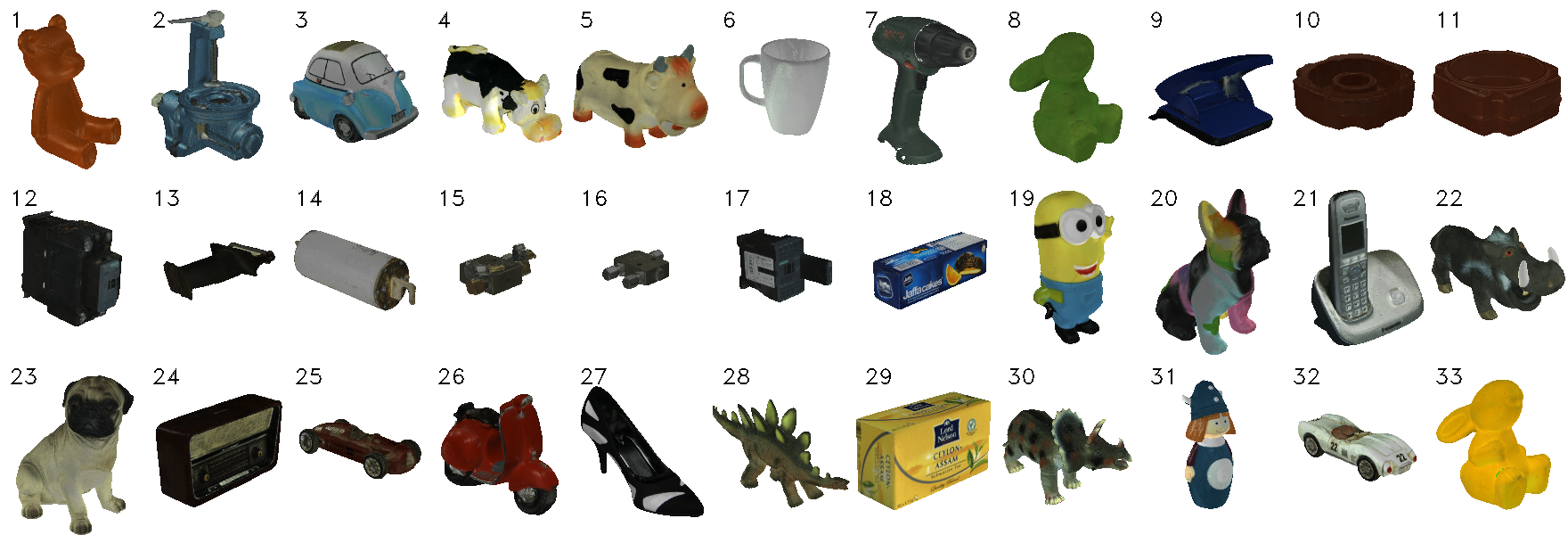}
	\caption{Rendered reconstructed 3D models of HomebrewedDB.}
	\label{fig:models}  
	\vspace{-0.8em}
\end{figure*}

Within our HomebrewedDB dataset we introduce the following:
\begin{enumerate}[noitemsep, leftmargin=4.2mm]
    \item {\bf 33 highly accurately reconstructed 3D models} of toys, household objects and low-textured industrial objects of sizes varying from 10.1 to 47.7 cm in diameter.
    \item {\bf 13 sequences}, each containing 1340 frames filmed using 2 different RGB-D sensors. Scenes span a range of complexity from simple (3 objects on a plain background) to complex (highly occluded with 8 objects and extensive clutter) (Fig.~\ref{fig:sample_scenes}). Also, two sequences feature drastic light changes or contain objects with altered textures (Fig.~\ref{fig:altered_scenes}).
    \item {\bf Precise 6D pose annotations} for dataset objects in the scenes, which were obtained using an automated pipeline (see Section~\ref{GTCreation}).
    \item {\bf A set of benchmarks} to facilitate comprehensive evaluation of object detection and 6D pose estimation methods.
\end{enumerate} 

The following sections detail the dataset creation, including calibration of RGB-D sensors, reconstruction of 3D models, depth correction, acquisition of image sequences and creation of ground truth annotations. We believe that the described steps would serve as a sufficient guide on how to extend an existing or create a new dataset for 6D pose estimation.

\subsection{Calibration of RGB-D Sensors}
For the footage of validation and test sequences we used two RGB-D sensors: the structured-light PrimeSense Carmine 1.09 and the time-of-flight Microsoft Kinect 2. Intrinsic and distortion parameters of both sensors were estimated during the calibration procedure. We used ArUco board~\cite{Garrido-JuradoMMM14}, which has yielded better calibration results in comparison with the classical checkerboard pattern and the corresponding intrinsic calibration module in OpenCV~\cite{opencv_library}. As a result of calibration, the root-mean squared reprojection error calculated at the corners of the ArUco markers is $\leq0.5$ and $\leq0.3$ px for Carmine and Kinect 2 respectively. Intrinsic and distortion parameters for both sensors are provided with the dataset. Because depth and color images were obtained from two different cameras, depth-to-color registration was performed using OpenNI 2.2 Driver for Carmine and Windows SDK 2.0 for Kinect 2. Given that the scenes were recorded independently with each of the sensors, there was no need in extrinsic calibration of the cameras.

\subsection{Sequence Acquisition}
In total we acquired 1 handheld and 2 turntable sequences for each of the scenes with each RGB-D sensor. Turntable sequences capture a full 360\degree rotation of a markerboard with objects on it using a camera mounted on a tripod. Each turntable sequence has 170 RGB and depth images filmed with elevation angles of 30\degree and 45\degree. Together with the ground truth 6D pose labels they form a validation dataset. In contrast, the test sequences were recorded in handheld mode. There were two major reasons for the handheld recording instead of using a controlled setup similar to those in T-LESS~\cite{hodan2017tless} or BigBIRD~\cite{SinghSNAA14}. The first clear advantage is the close resemblance to regular camera use, while the second one is the ability to introduce more variation to camera poses in terms of considerable scale changes as well as in-plane rotation. For the test sequences, a total of 1000 RGB and corresponding depth images of each scene were captured with each sensor.
\begin{figure*}[ht!]
   \begin{minipage}[b]{.19\linewidth}
	\includegraphics[width=1.0\linewidth]{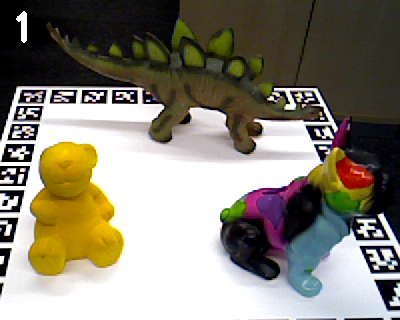}
  \end{minipage}
	\hfill
   \begin{minipage}[b]{.19\linewidth}
	\includegraphics[width=1.0\linewidth]{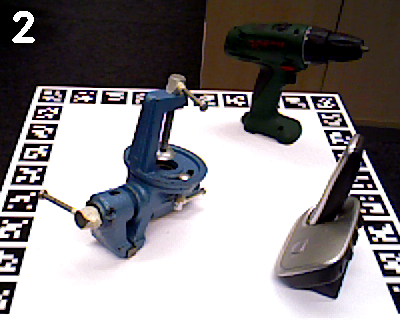}
  \end{minipage}
    \hfill
   \begin{minipage}[b]{.19\linewidth}
	\includegraphics[width=1.0\linewidth]{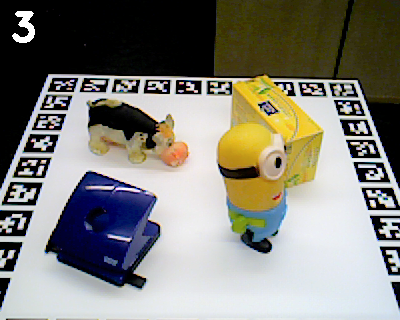}
  \end{minipage}
  \hfill
  \begin{minipage}[b]{.19\linewidth}
	\includegraphics[width=1.0\linewidth]{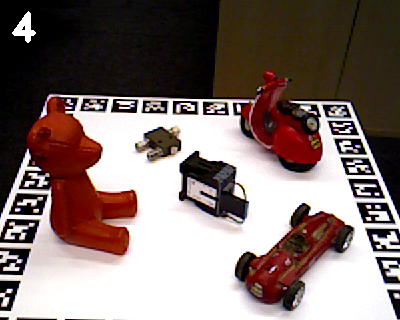}
  \end{minipage}
  	\hfill
	\begin{minipage}[b]{.19\linewidth}
	\includegraphics[width=1.0\linewidth]{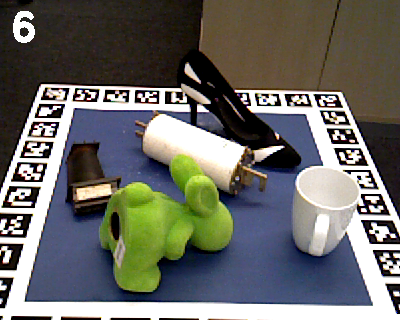}
  \end{minipage}
   \begin{minipage}[b]{.19\linewidth}
	\includegraphics[width=1.0\linewidth]{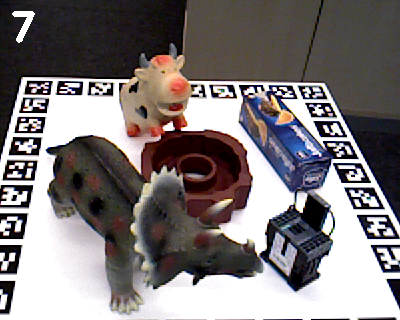}
  \end{minipage}
	\hfill
 \begin{minipage}[b]{.19\linewidth}
	\includegraphics[width=1.0\linewidth]{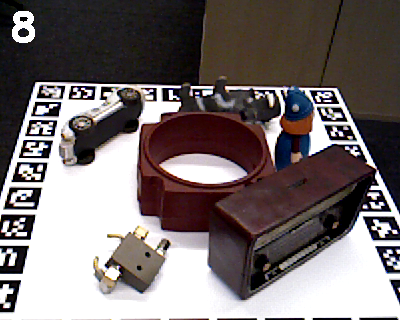}
  \end{minipage}
  	\hfill
   \begin{minipage}[b]{.19\linewidth}
	\includegraphics[width=1.0\linewidth]{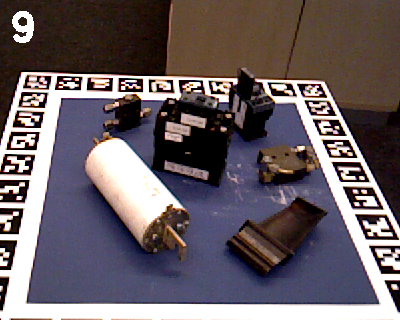}
  \end{minipage}
  \hfill
   \begin{minipage}[b]{.19\linewidth}
	\includegraphics[width=1.0\linewidth]{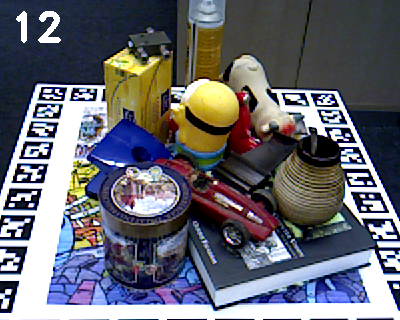}
  \end{minipage}
	\hfill
   \begin{minipage}[b]{.19\linewidth}
	\includegraphics[width=1.0\linewidth]{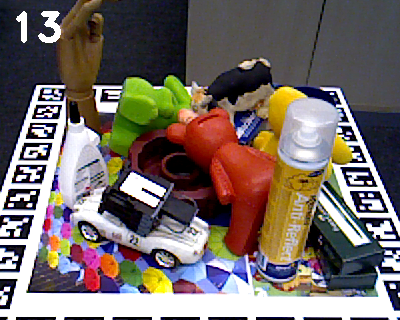}
  \end{minipage}

	\caption{Sample RGB images from the sequences presented in the HomebrewedDB dataset. Complexity of the scenes varies in terms of number and size of objects, levels of occlusion and clutter.}
	\label{fig:sample_scenes} 
	
\end{figure*}
\begin{figure*}[ht!]
\begin{minipage}[b]{0.32\linewidth}
	\includegraphics[width=1.0\linewidth]{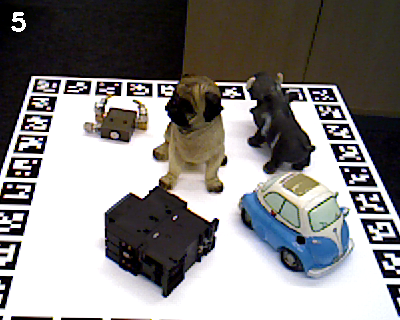}
  \end{minipage}
  	\hfill
	\begin{minipage}[b]{.32\linewidth}
	\includegraphics[width=1.0\linewidth]{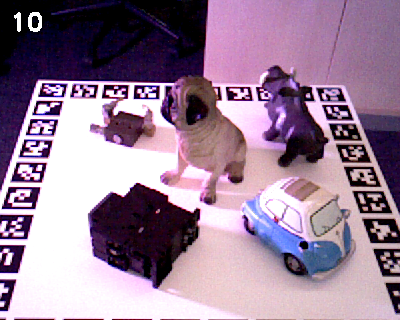}
  \end{minipage}
	\hfill
	\begin{minipage}[b]{0.32\linewidth}
	\includegraphics[width=1.0\linewidth]{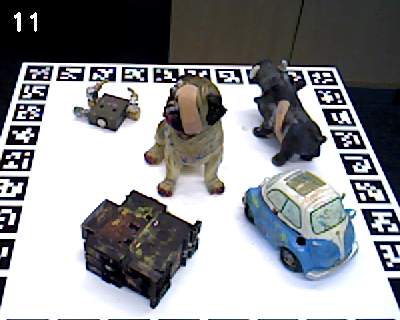}
  \end{minipage}
  
	\caption{Sample RGB images from sequences belonging to the domain adaptation benchmark.}
	\label{fig:altered_scenes} 
	\vspace{-0.8em}
\end{figure*}
While shooting each of those, a full pass around the markerboard was made. Within the test sequences, the distance from the camera to an object was varied from 0.42 to 1.43 m, while the elevations were within 11\degree-87\degree. Both color and depth images in the sequences that were recorded with the Carmine sensor have a default resolution of 640$\times$480 px, while Kinect 2 RGB images are of size 1920$\times$1080 px and depth images of 512$\times$424 px. The depth images were resized to the dimension of RGB images during registration.

For each of the scenes (Figs. ~\ref{fig:sample_scenes} and \ref{fig:altered_scenes}) target objects were placed on the markerboard with ArUco markers facilitating camera pose estimation. For the simpler scenarios, we placed the objects on a monochrome (white or dark-blue) Lambertian surface, and for the more complicated ones the objects resided on a multicolor reflective surface, being in some cases on top of each other. Also, more complicated sequences featured severe occlusions as well as objects not present in the dataset to make the scene more cluttered.

One new feature introduced in HomebrewedDB is a sequence aiming to test the robustness of detection and pose estimation methods with respect to significant changes in lighting conditions. To create it, we used a spotlight to repeatedly project series of light patterns with different colors and intensities onto the scene. We also created a domain adaptation sequence to evaluate robustness to considerable texture changes. For that we altered the textures of the objects by selectively painting some part of them with chalks of various colors.

% Besides, we created another sequence where we altered the textures of the objects by selectively painting some part of them with chalks of various colors. This sequence was created with a main goal to evaluate robustness to considerable texture changes.

\subsection{3D Model Reconstruction}
To obtain the 3D models presented in HomebrewedDB (Fig.~\ref{fig:models}), we first scanned each object from multiple viewpoints using Artec Eva~\cite{artec_eva}, a structured light 3D scanner. We opted for Artec Eva since it provides precise depth measurements and high resolution texture maps, which are crucial for reconstructing high-quality 3D models. The following pipeline for converting the scans to completely reconstructed 3D models proceeded using Artec Studio software. First, raw meshes were reconstructed for each of the scanned viewpoints. Secondly, we manually removed unnecessary parts of the meshes and then aligned them. Then we removed outliers and minor artifacts from the models. After that we proceeded with global optimization of the mesh structure, including inpainting minor holes in the model and as inducing smoothness of the mesh. Next, we back-projected high-resolution textures onto the resulting 3D model. Finally, we used MeshLab~\cite{meshlab} to center and and axis-align the models and computed surface normals as weighted sum of normals of the incident facets~\cite{facet_normals}.

\subsection{Depth Correction}
As have others ~\cite{sturm12iros, hodan2017tless}, we observed that depth measurements by both Carmine and Kinect 2 have systematic errors: we found that the measured depth values were always slightly different from those calculated from the markers in images captured with calibrated RGB cameras. Although it has been reported ~\cite{sturm12iros} that a single correction multiplier is sufficient to address the depth measurement error, we confirmed in our setup what Hodan \etal~\cite{hodan2017tless} had found - that first degree polynomial works better as a correction factor for the depth measurements in our setup. Using regularized least squares to account for noise in the measurements we derived the following linear depth-correction models:  $d_c = 1.0391 \cdot d - 15.8$ for Carmine and $d_c = 1.0186 \cdot d - 13.1$ for Kinect 2 (measured in millimeters). After applying the corrections, we found that the mean absolute difference from expected depth had been reduced from 14.7 mm to 2.03 mm and from 5.81 mm to 2.66 mm for Carmine and Kinect 2 respectively. We applied the corrections models to the entire dataset, so that no further user action is required.

\subsection{Creation of Ground Truth Annotations \label{GTCreation}}
The estimation of 6D ground truth objects poses for each frame in the sequence proceeded as follows. First, the markerboard pose was estimated, providing us with the camera trajectory around the scene. Then we obtained a dense 3D reconstruction by signed distance field fusion of depth maps of a scene with a method of Curless and Levoy~\cite{Curless96}, using all the images in a sequence. 

The next step was to estimate a rigid body transformation of a 3D model from its own coordinate system to the coordinate system of the scene (i.e., markerboard). Given that the locations of the objects with respect to the markerboard did not change when imaged with different sensors, it was sufficient to get the object pose in the markerboard coordinate system and then transfer it to a new sensor or camera coordinate system, thereby avoiding performing reconstruction for each new sensor. For the purpose of estimating 3D model poses in the reconstructed scene we used a method by Drost \etal~\cite{drost2010CVPR}, which is based on point-pair feature representation of a target model used for  local matching via an efficient voting scheme on a reduced 2D search space. Having a camera pose in each image estimated from the markerboard as well as having object poses in the markerboard coordinate system, 6D object poses for each of the frames can easily be computed. However, rendering the 3D models on top of RGB images revealed that even though the method by Drost \etal~\cite{drost2010CVPR} gives a good initial estimate of a pose, in many cases there remain visible discrepancies, which must be mitigated in the process of further refinement. 

To improve the poses, we opted for 2D edge-based ICP \cite{Drummond:2002:RVT:628329.628805} refinement. For each object in the sequence we automatically selected RGB images where the object was not occluded. This was done by rendering all the objects in the scene with the estimated initial poses and calculating the fraction of visible pixels for the target object. Multiview consistency was enforced such that all the camera poses stayed fixed, and optimization was only done for the object pose in the scene coordinate system. Edge-based refinement was performed on RGB images because depth measurements were relatively noisy and we observed minor misregistrations between depth and RGB images, particularly for those captured with Kinect 2.

\subsection{Accuracy of Ground Truth Poses}
To evaluate the accuracy of the computed ground truth poses, we followed the same procedure as introduced by Hodan \etal ~\cite{hodan2017tless}. We rendered the 3D objects using computed ground truth poses and for each pixel pair with valid depths in both rendered and captured images we computed the difference $\delta = d_c - d_r$, where $d_c$ and $d_r$ are captured and rendered depth values, respectively. The statistics obtained over the whole test set are presented in Tab.~\ref{table:gt_pose_accuracy}. As in T-LESS ~\cite{hodan2017tless}, differences exceeding 5 cm were omitted from the statistics as outliers. In HomebrewedDB, such measurements amounted to $3.7\%$, mostly caused by the clutter objects occluding the target objects, sensor measurement noise or minor discrepancies between reconstructed 3D models and real-world objects.

% \begin{table}[h]
% 	\begin{center}
% 		\begin{tabular}{|l|c|c|c|c|}
% 			\hline
% 			Sensor & $\mu_{\delta}$   & $\sigma_{\delta}$ & $\mu_{\left|\delta\right|}$ & $med_{\left|\delta\right|}$ \\
% 		    \hline
% 			%\midrule
% 			\shft           PrimeSense & 0.11 & 6.25 & 1.71 & 2.56\\
% 			\shft           Kinect 2 & 0.22 & 7.38 & 0.87 & 9.12\\
% 			\hline
% 		\end{tabular}
% 		\end{center}
% 	\caption{Differences between the depth of object rendered models at the ground truth poses and the captured depth
% (in mm). $\mu_{\delta}$ and $\sigma_{\delta}$ is the mean and the standard deviation
% of the differences, $\mu_{\left|\delta\right|}$ and $med_{\left|\delta\right|}$ is the mean and the median of the absolute differences}
% 	\label{table:gt_pose_accuracy}
% \end{table}\

\setlength{\tabcolsep}{15pt}
\begin{table}[t]
  \centering
  \resizebox{1\columnwidth}{!}{
    \begin{tabular}{r|cccc}
    \toprule
    \textbf{Sensor} & $\pmb{\mu_{\delta}}$   & $\pmb{\sigma_{\delta}}$ & $\pmb{\mu_{\left|\delta\right|}}$ & $\pmb{med_{\left|\delta\right|}}$  \\
    \midrule
    Carmine & 0.11 & 6.25 & 1.71 & 2.56 \\
    Kinect 2 & 0.22 & 7.38 & 0.87 & 9.12 \\
    \bottomrule
    \end{tabular}%
    }
    \caption{Differences between the depth of object rendered models at the ground truth poses and the captured depth (in mm). $\mu_{\delta}$ and $\sigma_{\delta}$ is the mean and the standard deviation of the differences, $\mu_{\left|\delta\right|}$ and $med_{\left|\delta\right|}$ is the mean and the median of the absolute differences.}
  \label{table:gt_pose_accuracy}%
    \vspace{-1em}
\end{table}%
From the presented results it can be seen that rendered depth maps align well with the depth maps obtained with Carmine, resulting in mean depth difference close to zero and absolute mean of differences $\leq 2$ mm. For Kinect 2 depth differences mean and absolute mean values also stay very close to zero: the absolute median value is notably higher than the absolute mean, signifying that the distribution of the depth differences is left-skewed. As noted in T-LESS ~\cite{hodan2017tless}, this might be caused by slight misregistration of RGB and depth images captured by Kinect 2, as well as higher magnitude of noise in the measurements of this depth sensor based on the time-of-flight principle.

%% file: content-experiments.tex
\setlength{\tabcolsep}{8pt}
\begin{table*}[t]
  \centering
  \resizebox{1\textwidth}{!}{%
    \begin{tabular}{c|r|ccccccccccc|ccc}
    \multicolumn{1}{c}{} &  &
    \multicolumn{10}{c}{\textbf{Per Scene}} &  &
    \multicolumn{3}{c}{\textbf{Texture / Illumination}} \\
    \midrule
    \multicolumn{1}{r}{} & \textbf{Scene ID} & 1 & 2 & 3 & 4 & 5 & 6 & 7 & 8 & 9 & 12 & 13 & 5 & 10 & 11 \\
    \midrule
    \multirow{3}[2]{*}{\begin{sideways}\textbf{Pose}\end{sideways}} & \textbf{ADD 10\%} & \cellcolor[rgb]{ .749,  .749,  .749}50.85 & \cellcolor[rgb]{ .784,  .784,  .784}45.08 & \cellcolor[rgb]{ .851,  .851,  .851}33.88 & \cellcolor[rgb]{ .886,  .886,  .886}27.25 & \cellcolor[rgb]{ .898,  .898,  .898}25.40 & \cellcolor[rgb]{ .933,  .933,  .933}19.19 & \cellcolor[rgb]{ .894,  .894,  .894}25.85 & \cellcolor[rgb]{ .98,  .98,  .98}11.08 & 7.60 & \cellcolor[rgb]{ .996,  .996,  .996}8.68 & \cellcolor[rgb]{ .969,  .969,  .969}13.36 & \cellcolor[rgb]{ .749,  .749,  .749}25.40 & 16.73 & 16.77 \\
      & \textbf{ADD 30\%} & \cellcolor[rgb]{ .749,  .749,  .749}81.34 & \cellcolor[rgb]{ .765,  .765,  .765}78.46 & \cellcolor[rgb]{ .82,  .82,  .82}64.53 & \cellcolor[rgb]{ .867,  .867,  .867}53.10 & \cellcolor[rgb]{ .863,  .863,  .863}54.00 & \cellcolor[rgb]{ .898,  .898,  .898}45.39 & \cellcolor[rgb]{ .922,  .922,  .922}39.99 & \cellcolor[rgb]{ .961,  .961,  .961}29.78 & 20.12 & \cellcolor[rgb]{ .98,  .98,  .98}25.68 & \cellcolor[rgb]{ .949,  .949,  .949}33.41 & \cellcolor[rgb]{ .749,  .749,  .749}54.00 & 40.32 & \cellcolor[rgb]{ .992,  .992,  .992}40.83 \\
      & \textbf{ADD 50\%} & \cellcolor[rgb]{ .749,  .749,  .749}88.71 & \cellcolor[rgb]{ .765,  .765,  .765}85.69 & \cellcolor[rgb]{ .808,  .808,  .808}75.46 & \cellcolor[rgb]{ .843,  .843,  .843}66.98 & \cellcolor[rgb]{ .863,  .863,  .863}61.95 & \cellcolor[rgb]{ .89,  .89,  .89}55.26 & \cellcolor[rgb]{ .929,  .929,  .929}46.76 & \cellcolor[rgb]{ .957,  .957,  .957}39.64 & 29.24 & \cellcolor[rgb]{ .98,  .98,  .98}34.49 & \cellcolor[rgb]{ .933,  .933,  .933}45.33 & \cellcolor[rgb]{ .749,  .749,  .749}61.95 & \cellcolor[rgb]{ .961,  .961,  .961}51.66 & 49.72 \\
    \midrule
    \multirow{3}[2]{*}{\begin{sideways}\textbf{Detect.}\end{sideways}} & \textbf{Precision} & \cellcolor[rgb]{ .749,  .749,  .749}0.79 & \cellcolor[rgb]{ .812,  .812,  .812}0.62 & \cellcolor[rgb]{ .761,  .761,  .761}0.76 & \cellcolor[rgb]{ .765,  .765,  .765}0.76 & \cellcolor[rgb]{ .804,  .804,  .804}0.65 & \cellcolor[rgb]{ .875,  .875,  .875}0.46 & \cellcolor[rgb]{ .792,  .792,  .792}0.68 & \cellcolor[rgb]{ .859,  .859,  .859}0.51 & \cellcolor[rgb]{ .953,  .953,  .953}0.26 & \cellcolor[rgb]{ .925,  .925,  .925}0.33 & 0.13 & \cellcolor[rgb]{ .749,  .749,  .749}0.65 & \cellcolor[rgb]{ .839,  .839,  .839}0.57 & 0.43 \\
      & \textbf{Recall} & \cellcolor[rgb]{ .749,  .749,  .749}0.95 & \cellcolor[rgb]{ .855,  .855,  .855}0.64 & \cellcolor[rgb]{ .796,  .796,  .796}0.82 & \cellcolor[rgb]{ .761,  .761,  .761}0.92 & \cellcolor[rgb]{ .8,  .8,  .8}0.80 & \cellcolor[rgb]{ .843,  .843,  .843}0.68 & \cellcolor[rgb]{ .788,  .788,  .788}0.84 & \cellcolor[rgb]{ .859,  .859,  .859}0.62 & \cellcolor[rgb]{ .957,  .957,  .957}0.33 & \cellcolor[rgb]{ .937,  .937,  .937}0.40 & 0.20 & \cellcolor[rgb]{ .749,  .749,  .749}0.80 & \cellcolor[rgb]{ .914,  .914,  .914}0.63 & 0.53 \\
      & \textbf{mAP} & \cellcolor[rgb]{ .749,  .749,  .749}0.82 & \cellcolor[rgb]{ .859,  .859,  .859}0.48 & \cellcolor[rgb]{ .784,  .784,  .784}0.72 & \cellcolor[rgb]{ .765,  .765,  .765}0.78 & \cellcolor[rgb]{ .812,  .812,  .812}0.64 & \cellcolor[rgb]{ .898,  .898,  .898}0.36 & \cellcolor[rgb]{ .812,  .812,  .812}0.64 & \cellcolor[rgb]{ .882,  .882,  .882}0.41 & \cellcolor[rgb]{ .973,  .973,  .973}0.14 & \cellcolor[rgb]{ .953,  .953,  .953}0.20 & 0.04 & \cellcolor[rgb]{ .749,  .749,  .749}0.64 & \cellcolor[rgb]{ .925,  .925,  .925}0.42 & 0.32 \\
    \bottomrule
    \end{tabular}%

    }
   \caption{Result of object detection and pose estimation presented on two benchmarks: (1) per scene benchmark spanning over 11 scenes of the dataset and the (2) domain adaptation benchmark evaluating the detector's generalization capabilities.}

  \label{tab:per_scene_dom_ad}%
  \vspace{-0.8em}
\end{table*}%

In this section, we present a set of benchmarks assessing the performance of a detector with respect to a variety of different conditions. In particular, such aspects as scalability and resistance to occlusions, different illumination conditions and changes in object texture are tested.

\subsection{Evaluation Metrics}
We use standard metrics for evaluating performance in 2D object detection: precision, recall and mean average precision (mAP). Conventionally, we consider an object to be correctly detected if the intersection over union (IoU) between the ground-truth and predicted bounding boxes is $\geq0.5$. To evaluate the correctness of the estimated 6D poses, as others have done ~\cite{hinterstoisserMultimodalTemplatesRealtime2011b,kehlSSD6DMakingRGBbased2017b,zakharov2019dpod}, we use the ADD score, which is defined as the average Euclidean distance between the model vertices transformed with ground truth and predicted poses:
\begin{equation}
\label{add_standard}
m = \avgunder_{\mathbf{x} \in \mathcal{M}} \norm{(\mathbf{R}\mathbf{x} + \mathbf{t}) - (\mathbf{\hat{R}}\mathbf{x} + \mathbf{\hat{t}})}_2,
\end{equation}
where  $\mathcal{M}$ is a set of vertices of a 3D model, ($\mathbf{R}, \mathbf{t}$) and ($\mathbf{\hat{R}},\mathbf{\hat{t}}$)  are ground truth and predicted rotation and translation, respectively.
As mentioned in ~\cite{hinterstoisserMultimodalTemplatesRealtime2011b}, a predicted pose is considered to be correct if ADD calculated with this pose is less than 10\% of a model diameter. However, in case of more complicated scenes, there is only a small fraction of poses falling into this category. Therefore, we also report ADD for the thresholds of 30\% and 50\% to give a broader overview of pose quality as well as to give an estimate of proportion of the poses which could be still a subject for further refinement.  

%\sergey{Here we describe ADD, Precision, Recall, MAP.}

\subsection{Dense Object Pose Detector}
For the performance evaluation we opted for the recently introduced Dense Pose Object Detector (DPOD)~\cite{zakharov2019dpod}, owing to its excellent performance on LineMOD and OCCLUSION datasets and its ability to be trained on both real and synthetic data. Other detectors also could be used, and we highly encourage prospective researchers to perform evaluation on HomebrewedDB in order to facilitate progress in the direction of scalability and training from CAD models. 

The DPOD detector is based on DensePose~\cite{Guler2018DensePose} applied to human pose detection and formalizes the object detection problem as a dense correspondence estimation problem. From the input image DPOD outputs the multi-labeled object segmentation mask and UV correspondence map. Obtained correspondences are further used as input to PnP and RANSAC for pose estimation. The more correspondences obtained, the less prone the detector is to wrong matches, and the easier and more accurate pose estimation becomes.
%UV map has been automatically created by spherical or cylindrical projection during training for each object. Since the values of UV map are limited to a discrete range, it is much easier to regress them than 3D object coordinates~\cite{OJafari17_iPose}. UV coordinates represent a direct mapping between the input image and vertices of 3D object model. Therefore, 2D-3D correspondences are directly obtained and can be further used as an input to  PnP and RANSAC for pose estimation. Large number of correspondences makes the detector less prone to wrong matches and allows easier and more accurate pose estimation.  

\subsection{Scalability Benchmark}
The first benchmark was introduced with a goal to evaluate the method's scalability with respect to the number of objects. The main requirement is to train a single network for all the objects available in the dataset and test it on a set of sequences containing all the objects. Specifically, for this purpose we jointly evaluate a method on the sequences from 1 to 8 (inclusive), which form a minimal subset of sequences with all 33 objects present. For this benchmark object detection and pose estimation results are reported separately for each object.

From the results presented in Tab.~\ref{tab:scalability} it can be seen that the best detection and pose estimation performance is achieved for bigger objects with distinct textures and geometric features (e.g., 28, 30), while detection of smaller low-textured or glossy industrial objects (e.g., 12, 13, 14)  is a considerable challenge for DPOD. Besides, pose estimation results demonstrate that the DPOD detector does not scale particularly well for this task: for 17 out of 33 objects, ADD with 10\% threshold is under 10\%, and there are no instances for which the achieved score was higher than 50\%.

\subsection{Scene Benchmarks}
Our dataset presents a collection of 13 scenes with a varying degree of difficulty, and each of these scenes represents a separate benchmark, where all the objects it contains are used for training. Except for the scenes with altered illumination conditions or texture changes (i.e., 10 and 11), we include all the scenes in the per-scene benchmarks. All the object detection and pose estimation scores are averaged over all the objects present in the scene. In Tab.~\ref{tab:per_scene_dom_ad} the objects detection and pose estimation performance per scene is presented. As expected, DPOD demonstrates significantly better performance if both tasks in the sequences with the smaller number of objects, as well as less significant occlusions and no clutter. Also, low scores in both detection and pose estimation are reported for scene 8, which is composed exclusively of industrial objects.        

\subsection{Domain Adaptation Benchmark}
The main goal of introducing the domain adaptation benchmark is to test the robustness of a method to significant changes in lighting conditions and objects textures. It is composed of three scenes (5, 10 and 11) with the same set of objects, but differing in terms of illumination and the color of object surfaces. Scene 5 was captured in ambient lighting conditions with no alteration of the objects' appearance. In contrast, in scene 10 we used a spotlight to project light of different colors and intensity onto the imaged objects to introduce considerable variations in illumination, whereas in scene 11, we applied paint on the objects' surfaces to alter their texture. For this benchmark, object detection and 6D pose estimation scores are presented per scene.

As can be seen from results in Tab.~\ref{tab:per_scene_dom_ad}, performance in both detection and pose estimation is notably better in the cases of no added illumination or texture changes. Under normal conditions the resulting poses are 37\% more accurate based on ADD with a 10\%  threshold when compared to those under altered conditions. Also, object detection performance falls far behind under altered conditions compared to normal conditions, resulting in 46\% and 59\% lower mAP scores for the scenes with variations in light and texture, respectively. These results suggest that there is a lot of room for further adaptation of the DPOD detector to changing environments.

\setlength{\tabcolsep}{6pt}
\begin{table}[t]
  \centering
  \resizebox{1\columnwidth}{!}{%
    \begin{tabular}{c|cccccc}
    \textbf{Obj. ID} & \textbf{ADD 10\%} & \textbf{ADD 30\%} & \textbf{ADD 50\%} & \textbf{Precision} & \textbf{Recall} & \textbf{mAP} \\
    \midrule
    1 & \cellcolor[rgb]{ .835,  .835,  .835}23.04 & \cellcolor[rgb]{ .733,  .733,  .733}68.30 & \cellcolor[rgb]{ .71,  .71,  .71}81.04 & \cellcolor[rgb]{ .69,  .69,  .69}0.86 & \cellcolor[rgb]{ .655,  .655,  .655}0.98 & \cellcolor[rgb]{ .69,  .69,  .69}0.85 \\
    2 & \cellcolor[rgb]{ .875,  .875,  .875}17.85 &c \cellcolor[rgb]{ .776,  .776,  .776}58.02 & \cellcolor[rgb]{ .729,  .729,  .729}75.87 & \cellcolor[rgb]{ .8,  .8,  .8}0.62 & \cellcolor[rgb]{ .694,  .694,  .694}0.89 & \cellcolor[rgb]{ .8,  .8,  .8}0.56 \\
    3 & \cellcolor[rgb]{ .902,  .902,  .902}14.18 & \cellcolor[rgb]{ .788,  .788,  .788}55.41 & \cellcolor[rgb]{ .745,  .745,  .745}72.40 & \cellcolor[rgb]{ .663,  .663,  .663}0.93 & \cellcolor[rgb]{ .678,  .678,  .678}0.92 & \cellcolor[rgb]{ .678,  .678,  .678}0.88 \\
    4 & \cellcolor[rgb]{ .937,  .937,  .937}9.09 & \cellcolor[rgb]{ .863,  .863,  .863}36.74 & \cellcolor[rgb]{ .808,  .808,  .808}55.43 & \cellcolor[rgb]{ .788,  .788,  .788}0.66 & \cellcolor[rgb]{ .733,  .733,  .733}0.79 & \cellcolor[rgb]{ .812,  .812,  .812}0.53 \\
    5 & \cellcolor[rgb]{ .918,  .918,  .918}11.70 & \cellcolor[rgb]{ .835,  .835,  .835}43.23 & \cellcolor[rgb]{ .796,  .796,  .796}58.60 & \cellcolor[rgb]{ .663,  .663,  .663}0.93 & \cellcolor[rgb]{ .655,  .655,  .655}0.98 & \cellcolor[rgb]{ .667,  .667,  .667}0.91 \\
    6 & 0.00 & \cellcolor[rgb]{ .996,  .996,  .996}3.51 & \cellcolor[rgb]{ .973,  .973,  .973}12.13 & \cellcolor[rgb]{ .78,  .78,  .78}0.67 & \cellcolor[rgb]{ .78,  .78,  .78}0.68 & \cellcolor[rgb]{ .839,  .839,  .839}0.46 \\
    7 & \cellcolor[rgb]{ .894,  .894,  .894}15.15 & \cellcolor[rgb]{ .831,  .831,  .831}44.12 & \cellcolor[rgb]{ .788,  .788,  .788}60.74 & \cellcolor[rgb]{ .784,  .784,  .784}0.66 & \cellcolor[rgb]{ .78,  .78,  .78}0.68 & \cellcolor[rgb]{ .843,  .843,  .843}0.45 \\
    8 & \cellcolor[rgb]{ .867,  .867,  .867}18.67 & \cellcolor[rgb]{ .839,  .839,  .839}42.74 & \cellcolor[rgb]{ .82,  .82,  .82}52.28 & \cellcolor[rgb]{ .937,  .937,  .937}0.33 & \cellcolor[rgb]{ .965,  .965,  .965}0.24 & \cellcolor[rgb]{ .98,  .98,  .98}0.08 \\
    9 & \cellcolor[rgb]{ .984,  .984,  .984}2.26 & \cellcolor[rgb]{ .953,  .953,  .953}13.71 & \cellcolor[rgb]{ .914,  .914,  .914}27.58 & \cellcolor[rgb]{ .812,  .812,  .812}0.60 & \cellcolor[rgb]{ .808,  .808,  .808}0.62 & \cellcolor[rgb]{ .871,  .871,  .871}0.37 \\
    10 & \cellcolor[rgb]{ .98,  .98,  .98}3.15 & \cellcolor[rgb]{ .929,  .929,  .929}19.63 & \cellcolor[rgb]{ .902,  .902,  .902}30.96 & \cellcolor[rgb]{ .678,  .678,  .678}0.89 & \cellcolor[rgb]{ .71,  .71,  .71}0.86 & \cellcolor[rgb]{ .722,  .722,  .722}0.76 \\
    11 & \cellcolor[rgb]{ .996,  .996,  .996}0.71 & \cellcolor[rgb]{ .992,  .992,  .992}4.07 & \cellcolor[rgb]{ .98,  .98,  .98}9.87 & \cellcolor[rgb]{ .714,  .714,  .714}0.82 & \cellcolor[rgb]{ .655,  .655,  .655}0.98 & \cellcolor[rgb]{ .71,  .71,  .71}0.80 \\
    12 & 0.50 & \cellcolor[rgb]{ .996,  .996,  .996}3.74 & \cellcolor[rgb]{ .988,  .988,  .988}7.48 & \cellcolor[rgb]{ .875,  .875,  .875}0.47 & \cellcolor[rgb]{ .898,  .898,  .898}0.40 & \cellcolor[rgb]{ .941,  .941,  .941}0.19 \\
    13 & \cellcolor[rgb]{ .984,  .984,  .984}2.25 & \cellcolor[rgb]{ .929,  .929,  .929}20.22 & \cellcolor[rgb]{ .882,  .882,  .882}35.63 & \cellcolor[rgb]{ .824,  .824,  .824}0.58 & \cellcolor[rgb]{ .804,  .804,  .804}0.62 & \cellcolor[rgb]{ .875,  .875,  .875}0.36 \\
    14 & \cellcolor[rgb]{ .98,  .98,  .98}2.88 & \cellcolor[rgb]{ .914,  .914,  .914}24.04 & \cellcolor[rgb]{ .867,  .867,  .867}40.19 & \cellcolor[rgb]{ .902,  .902,  .902}0.41 & \cellcolor[rgb]{ .847,  .847,  .847}0.52 & \cellcolor[rgb]{ .933,  .933,  .933}0.21 \\
    15 & 0.00 & 1.88 & \cellcolor[rgb]{ .996,  .996,  .996}5.52 & \cellcolor[rgb]{ .937,  .937,  .937}0.33 & \cellcolor[rgb]{ .898,  .898,  .898}0.40 & \cellcolor[rgb]{ .961,  .961,  .961}0.14 \\
    16 & 0.00 & \cellcolor[rgb]{ .992,  .992,  .992}4.70 & \cellcolor[rgb]{ .98,  .98,  .98}9.94 & \cellcolor[rgb]{ .871,  .871,  .871}0.48 & \cellcolor[rgb]{ .914,  .914,  .914}0.36 & \cellcolor[rgb]{ .949,  .949,  .949}0.17 \\
    17 & \cellcolor[rgb]{ .996,  .996,  .996}0.72 & \cellcolor[rgb]{ .996,  .996,  .996}3.37 & \cellcolor[rgb]{ .953,  .953,  .953}17.35 & \cellcolor[rgb]{ .8,  .8,  .8}0.63 & \cellcolor[rgb]{ .89,  .89,  .89}0.42 & \cellcolor[rgb]{ .914,  .914,  .914}0.26 \\
    18 & 0.16 & 2.24 & 4.32 & \cellcolor[rgb]{ .867,  .867,  .867}0.49 & \cellcolor[rgb]{ .804,  .804,  .804}0.63 & \cellcolor[rgb]{ .89,  .89,  .89}0.32 \\
    19 & \cellcolor[rgb]{ .945,  .945,  .945}7.70 & \cellcolor[rgb]{ .894,  .894,  .894}28.63 & \cellcolor[rgb]{ .843,  .843,  .843}45.88 & \cellcolor[rgb]{ .659,  .659,  .659}0.94 & \cellcolor[rgb]{ .682,  .682,  .682}0.92 & \cellcolor[rgb]{ .675,  .675,  .675}0.89 \\
    20 & \cellcolor[rgb]{ .835,  .835,  .835}23.22 & \cellcolor[rgb]{ .753,  .753,  .753}63.60 & \cellcolor[rgb]{ .729,  .729,  .729}75.73 & \cellcolor[rgb]{ .667,  .667,  .667}0.92 & \cellcolor[rgb]{ .667,  .667,  .667}0.96 & \cellcolor[rgb]{ .671,  .671,  .671}0.91 \\
    21 & \cellcolor[rgb]{ .941,  .941,  .941}8.37 & \cellcolor[rgb]{ .863,  .863,  .863}36.12 & \cellcolor[rgb]{ .827,  .827,  .827}50.66 & \cellcolor[rgb]{ .91,  .91,  .91}0.39 & \cellcolor[rgb]{ .969,  .969,  .969}0.23 & \cellcolor[rgb]{ .98,  .98,  .98}0.09 \\
    22 & \cellcolor[rgb]{ .929,  .929,  .929}10.04 & \cellcolor[rgb]{ .867,  .867,  .867}35.97 & \cellcolor[rgb]{ .808,  .808,  .808}55.02 & \cellcolor[rgb]{ .741,  .741,  .741}0.76 & \cellcolor[rgb]{ .741,  .741,  .741}0.78 & \cellcolor[rgb]{ .788,  .788,  .788}0.59 \\
    23 & \cellcolor[rgb]{ .886,  .886,  .886}16.18 & \cellcolor[rgb]{ .757,  .757,  .757}62.49 & \cellcolor[rgb]{ .706,  .706,  .706}82.20 & \cellcolor[rgb]{ .651,  .651,  .651}0.95 & \cellcolor[rgb]{ .651,  .651,  .651}0.99 & \cellcolor[rgb]{ .651,  .651,  .651}0.95 \\
    24 & \cellcolor[rgb]{ .973,  .973,  .973}4.08 & \cellcolor[rgb]{ .89,  .89,  .89}29.25 & \cellcolor[rgb]{ .808,  .808,  .808}55.78 & 0.19 & 0.15 & 0.03 \\
    25 & \cellcolor[rgb]{ .933,  .933,  .933}9.76 & \cellcolor[rgb]{ .851,  .851,  .851}39.50 & \cellcolor[rgb]{ .827,  .827,  .827}50.62 & \cellcolor[rgb]{ .745,  .745,  .745}0.75 & \cellcolor[rgb]{ .698,  .698,  .698}0.88 & \cellcolor[rgb]{ .761,  .761,  .761}0.66 \\
    26 & \cellcolor[rgb]{ .894,  .894,  .894}15.01 & \cellcolor[rgb]{ .8,  .8,  .8}51.83 & \cellcolor[rgb]{ .757,  .757,  .757}68.98 & \cellcolor[rgb]{ .729,  .729,  .729}0.78 & \cellcolor[rgb]{ .733,  .733,  .733}0.79 & \cellcolor[rgb]{ .773,  .773,  .773}0.63 \\
    27 & \cellcolor[rgb]{ .906,  .906,  .906}13.47 & \cellcolor[rgb]{ .8,  .8,  .8}51.78 & \cellcolor[rgb]{ .749,  .749,  .749}71.47 & \cellcolor[rgb]{ .773,  .773,  .773}0.68 & \cellcolor[rgb]{ .749,  .749,  .749}0.76 & \cellcolor[rgb]{ .816,  .816,  .816}0.52 \\
    28 & \cellcolor[rgb]{ .816,  .816,  .816}26.17 & \cellcolor[rgb]{ .757,  .757,  .757}63.30 & \cellcolor[rgb]{ .722,  .722,  .722}78.07 & \cellcolor[rgb]{ .722,  .722,  .722}0.80 & \cellcolor[rgb]{ .682,  .682,  .682}0.92 & \cellcolor[rgb]{ .733,  .733,  .733}0.73 \\
    29 & \cellcolor[rgb]{ .902,  .902,  .902}13.97 & \cellcolor[rgb]{ .855,  .855,  .855}38.59 & \cellcolor[rgb]{ .8,  .8,  .8}57.68 & \cellcolor[rgb]{ .659,  .659,  .659}0.93 & \cellcolor[rgb]{ .675,  .675,  .675}0.94 & \cellcolor[rgb]{ .678,  .678,  .678}0.89 \\
    30 & \cellcolor[rgb]{ .651,  .651,  .651}48.63 & \cellcolor[rgb]{ .651,  .651,  .651}88.71 & \cellcolor[rgb]{ .651,  .651,  .651}96.34 & \cellcolor[rgb]{ .718,  .718,  .718}0.80 & \cellcolor[rgb]{ .655,  .655,  .655}0.98 & \cellcolor[rgb]{ .714,  .714,  .714}0.79 \\
    31 & \cellcolor[rgb]{ .957,  .957,  .957}6.43 & \cellcolor[rgb]{ .922,  .922,  .922}21.85 & \cellcolor[rgb]{ .878,  .878,  .878}36.68 & \cellcolor[rgb]{ .706,  .706,  .706}0.83 & \cellcolor[rgb]{ .71,  .71,  .71}0.86 & \cellcolor[rgb]{ .733,  .733,  .733}0.74 \\
    32 & 0.00 & \cellcolor[rgb]{ .984,  .984,  .984}5.95 & \cellcolor[rgb]{ .976,  .976,  .976}10.71 & \cellcolor[rgb]{ .984,  .984,  .984}0.23 & \cellcolor[rgb]{ .992,  .992,  .992}0.17 & 0.04 \\
    33 & \cellcolor[rgb]{ .922,  .922,  .922}11.40 & \cellcolor[rgb]{ .827,  .827,  .827}45.69 & \cellcolor[rgb]{ .773,  .773,  .773}64.99 & \cellcolor[rgb]{ .659,  .659,  .659}0.93 & \cellcolor[rgb]{ .659,  .659,  .659}0.97 & \cellcolor[rgb]{ .667,  .667,  .667}0.91 \\
    \end{tabular}%

    }
      \caption{Results of object detection and pose estimation on scalability benchmark.}
    \vspace{-1em}
  \label{tab:scalability}%
\end{table}%

\subsection{Drawbacks of Training on Real Data}
The main reason we exclusively use synthetic data for training in our benchmarks comes from the inability of the detectors trained on a small-scale set of real data to generalize to different environments, which may differ in background, illumination, texture and other scene characteristics. 

To support our claim, we selected 2 recent state of the art detectors, YOLO6D~\cite{tekinRealTimeSeamlessSingle2017} and DPOD~\cite{zakharov2019dpod}, trained on real LineMOD data and tested them on our new sequence containing 3 LineMOD objects - a benchvise, a drill and a phone (Scene 2 in Fig.~\ref{fig:sample_scenes}). This sequence contains a minimal number of occlusions and no clutter; it may be regarded as one of the simplest scenes in the dataset. Besides, this sequence was captured with the same camera as LineMOD sequences (PrimeSense Carmine 1.09), making it similar in terms of color scheme and noise. Performance comparison of both these detectors on LineMOD sequence and our sequence can be seen in the "Real" section of Tab.~\ref{tab:real_lm_vs_our}. In spite of demonstrating very good performance when tested on sequences from LineMOD, when run on our sequence with LineMOD objects, both detectors experience a significant drop in pose estimation accuracy in terms of the ADD 10\% metric for all the objects. Specifically, the pose estimation accuracy of DPOD~\cite{zakharov2019dpod} dropped more than 2 times for the phone, while for YOLO6D~\cite{zakharov2019dpod} there were nearly no correctly predicted poses for the drill. 

As the second part of the experiment, we evaluated 2 detectors designed for training on synthetic data, DPOD~\cite{zakharov2019dpod} and SSD6D with model-based refinement~\cite{manhardtDeepModelBased6D}, on the same new sequence with LineMOD objects. From the results presented in the "Synthetic" section of Tab.~\ref{tab:real_lm_vs_our} one can see that there is no significant difference between the results obtained on LineMOD and HomebrewedDB sequences, meaning that there is no overfitting to a particular dataset.  In 5 out of 6 cases, the detectors demonstrated better performance on a simpler HomebrewedDB sequence, thus showing more predictable behavior than in the case of training on real data. Moreover, when trained on synthetic data, DPOD~\cite{zakharov2019dpod} can boast higher ADD 10\% scores for all the 3 objects in the HomebrewedDB test sequence. This fact once again supports the claim that training detectors on synthetic data leads to better generalization, in contrast to training on real data, when even a seemingly insignificant changes in the environment turn out to be a decisive factor leading to a significant decrease in pose estimation accuracy.

\setlength{\tabcolsep}{10pt}
\begin{table}[t]
  \centering
  \resizebox{1\columnwidth}{!}{%
    \begin{tabular}{c|c|c|ccc}
    & \textbf{Dataset} & \textbf{Method} & \textbf{Benchvise} & \textbf{Phone} & \textbf{Driller} \\
    \toprule
    \multirow{4}[4]{*}{\begin{sideways}\textbf{Real}\end{sideways}} & \multirow{2}[1]{*}{LM} & YOLO6D~\cite{tekinRealTimeSeamlessSingle2017} & 81.80 & 47.74 & 63.51 \\
          &       & DPOD~\cite{zakharov2019dpod} & 95.34 & 74.24 & 97.72 \\
\cmidrule{2-6}          & \multirow{2}[2]{*}{HB} & YOLO6D ~\cite{tekinRealTimeSeamlessSingle2017}& 15.30 & 6.50  & 0.10 \\
          &       & DPOD~\cite{zakharov2019dpod}  & 57.24 & 33.09 & 62.82 \\
    \midrule
    \multirow{4}[4]{*}{\begin{sideways}\textbf{Synthetic}\end{sideways}} & \multirow{2}[2]{*}{LM} & SSD6D + Ref.~\cite{manhardtDeepModelBased6D} & 44.30 & 26.20 & 26.90 \\
          &       & DPOD~\cite{zakharov2019dpod}  & 66.76 & 29.08 & 66.60 \\
\cmidrule{2-6}          & \multirow{2}[2]{*}{HB} & SSD6D + Ref.~\cite{manhardtDeepModelBased6D} & 59.40 & 29.30 & 25.10 \\
          &       & DPOD~\cite{zakharov2019dpod}  & 70.89 & 35.56 & 66.42 \\
    \bottomrule
    \end{tabular}%
    }
      \caption{Pose estimation results in terms of ADD 10\% metric on LineMOD sequences (LM) and HomebrewedDB (HB) sequence with the same objects.   }
  \label{tab:real_lm_vs_our}%
   \vspace{-1em}
\end{table}%

%% file: content-conclusion.tex
% !TeX spellcheck = en_US

In this work, we have presented a new challenging dataset for 6D object detection, covering the most important properties a solid object detector should have, namely scalability with respect to the number of objects and robustness to occlusions, illumination and appearance changes. This new dataset contains 33 objects spanning 13 scenes of various difficulty. To be able to compare the detectors to each other, we defined a set of benchmarks that test all of the above mentioned properties. Finally, we developed and presented a comparably simple, yet robust and fully automated pipeline that we used to build our dataset. We hope it will allow other researchers to be able to create their own datasets and thus promote research in 6D pose estimation.

%We plan to release the code for the other researchers to be able to create their own datasets and thus promote the research in 6D pose estimation. 